\documentclass{article}

    \usepackage[preprint]{neurips_2025}

\usepackage[utf8]{inputenc} 
\usepackage[T1]{fontenc}    
\usepackage{hyperref}       
\usepackage{url}            
\usepackage{booktabs}       
\usepackage{nicefrac}       
\usepackage{microtype}      
\usepackage{xcolor}         

\usepackage[small,compact]{titlesec}
\usepackage{amsmath,amssymb,amsfonts,mathabx,setspace}
\usepackage{graphicx,caption,epsfig,epsfig,wrapfig}
\usepackage{caption}
\usepackage[caption=false]{subfig}
\usepackage{float}
\usepackage{url,color,verbatim,algorithmicx,xcolor}
\usepackage[noend]{algpseudocode}
\usepackage[ruled,boxed]{algorithm} 

\usepackage{enumitem}

\def\R{\mathbb{R}}                            

\def\E{\mathbb{E}}

\newcommand{\innerp}[1]{\langle{#1}\rangle}

\newcommand{\argmin}[1]{\underset{#1}{\operatorname{arg}\operatorname{min}}\;}

\definecolor{darkmagenta}{rgb}{0.55, 0.0, 0.55}
\colorlet{tbd}{darkmagenta}

\newtheorem{theorem}{Theorem}

\newtheorem{corollary}[theorem]{Corollary}
\newtheorem{definition}[theorem]{Definition}

\newtheorem{lemma}[theorem]{Lemma}
\newtheorem{proposition}[theorem]{Proposition}
\newtheorem{remark}[theorem]{Remark}
\newenvironment{proof}[1][Proof]{\noindent\textbf{#1.} }{\ \rule{0.5em}{0.5em}}
\numberwithin{equation}{section}
\numberwithin{theorem}{section}

\title{
Transformer learns the cross-task prior and regularization for in-context learning
}

\author{%
Fei Lu \\
  Department of Mathematics, \\
  Johns Hopkins University, \\
  Baltimore, MD 21218, USA \\
  \texttt{feilu@math.jhu.edu}\\
\And
 Yue Yu\thanks{Corresponding Author} \\
  Department of Mathematics, \\
  Lehigh University, \\
  Bethlehem, PA 18015, USA \\
  \texttt{yuy214@lehigh.edu} \\
}

\begin{document}

\maketitle

\begin{abstract}
 Transformers have shown a remarkable ability for in-context learning (ICL), making predictions based on contextual examples. However, while theoretical analyses have explored this prediction capability, the nature of the inferred context and its utility for downstream predictions remain open questions. This paper aims to address these questions by examining ICL for inverse linear regression (ILR), where context inference can be characterized by unsupervised learning of underlying weight vectors. Focusing on the challenging scenario of rank-deficient inverse problems, where context length is smaller than the number of unknowns in the weight vectors and regularization is necessary, we introduce a linear transformer to learn the inverse mapping from contextual examples to the underlying weight vector. Our findings reveal that the transformer implicitly learns both a prior distribution and an effective regularization strategy, outperforming traditional ridge regression and regularization methods. A key insight is the necessity of low task dimensionality relative to the context length for successful learning. Furthermore, we numerically verify that the error of the transformer estimator scales linearly with the noise level, the ratio of task dimension to context length, and the condition number of the input data. These results not only demonstrate the potential of transformers for solving ill-posed inverse problems, but also provide a new perspective towards understanding the knowledge extraction mechanism within transformers.
\end{abstract}


\section{Introduction}\label{sec:intro}
 Transformers have shown a remarkable ability for in-context learning (ICL), once trained, they can make predictions based on contextual examples in new tasks without any parameter updates \cite{Brown2020Language,dai2023can,wies2023learnability,tian24joma}. While their predictive power is well-documented, especially in natural language processing and for forward prediction tasks, a deeper understanding of what transformers infer from the context and how this inferred knowledge is utilized remains an active area of research \cite{akyurek23learning,von2023transformers,bai2023transformers,oko2024pretrained,xie2022explanation,bhattamishra2024understanding,pathak24transformers,havrilla2024understanding,Reddy2024mechanistic}. Key open questions revolve around the nature of the learning and whether transformers can effectively tackle more complex tasks, such as those arising in scientific modeling and inverse problems. 

This paper aims to shed light on these questions by investigating ICL in the specific, yet fundamental, context of Inverse Linear Regression (ILR). In ILR, the objective is not merely to predict future outputs, but to infer the underlying linear relationship, the weight vector $w$, that generates the observed data. This setting provides a clear framework in which the inferred context can be characterized as an estimate of the underlying weight vector. We focus on a particularly challenging and practical scenario: rank-deficient inverse problems, where the context length is smaller than the dimensionality of the unknown weight vector. Such problems are inherently ill-posed and necessitate effective regularization to yield meaningful solutions.

Our central hypothesis is that a transformer, when trained to perform ICL for ILR across multiple related tasks, learns more than just a superficial pattern-matching heuristic. We posit that it learns an effective inverse mapping from contextual examples to the underlying weight vector by implicitly discovering both a prior distribution over these weight vectors and an appropriate regularization strategy. By successfully navigating these ill-posed problems, the transformer offers a new lens through which to understand its capacity for knowledge extraction and application. This study explores these capabilities, analyzing how a linear transformer learns this prior and regularization, and how its performance is influenced by the inherent low-dimensionality of the tasks.

\begin{table}[t]
  \centering
  \caption{Comparing in-context learning for forward and inverse linear regression, and ridge regression.}
  \label{tab:paradigm_comparison}
  \begin{tabular}{@{}lllp{5cm}p{4cm}@{}}
    \toprule
    \textbf{Paradigm} & \textbf{Context} 
    & \textbf{Goal} & \textbf{Notes} \\
    \midrule
    ICL–LR
      & $(x_{1:n},\,y_{1:n})$
      & $\displaystyle \hat y_{n+1}
        = f_\theta(x_{n+1}\mid x_{1:n},y_{1:n})$
      & $n\geq d$\,, cross‐task ICL ($\theta$-training);  no weight updates \\
    \addlinespace
    ICL–ILR 
      & $(x_{1:n},\,y_{1:n})$
      & $\displaystyle \widehat w
        = w_\theta(x_{1:n},y_{1:n})$
      & $n<d$\,, cross‐task ICL ($\theta$-training); no weight updates \\
    \addlinespace
    Ridge ILR
      & $(x_{1:n},\,y_{1:n})$
      & $\displaystyle \widehat w
        = (X^\top X + \lambda I_d)^{-1} X^\top Y$
      & no cross‐task sharing; $\lambda$ tuned per‐context \\
    \bottomrule
  \end{tabular}
\end{table}
\subsection{Problem formulation and main results}
\label{sec:problem}
We investigate In-Context Learning (ICL) for the Inverse Linear Regression (ILR) task. Given a \emph{context} consisting of $n$ input-output pairs $(x_{1:n}, y_{1:n}) = (x_1, y_1, \ldots, x_n, y_n)$, where $y_i = \langle x_i, w \rangle + \varepsilon_i$, the objective is to estimate the underlying linear relation $w \in \mathbb{R}^{d \times 1}$. Each context $(x_i, y_i)_{i=1}^n$ comprises i.i.d. samples from a joint distribution $P(x,y|w)$ specific to that $w$.

This ILR task fundamentally differs from standard ICL for linear regression, where the primary goal is to predict a new output $y_{n+1}$ given $x_{n+1}$ and the context. Instead, ICL-ILR focuses on the inverse problem: learning the underlying rule $w$ itself from the provided examples. This estimated $\widehat{w}$ can then facilitate interpretable insights or be used for subsequent predictions.

Our study addresses the challenging rank-deficient scenario where the context length $n$ is smaller than the ambient dimension $d$ of $w$ ($n<d$). In this scenario, the ILR is under-determined, necessitating a regularization. The core aim is to train a model that learns an inverse mapping that transforms a context to a $w$ estimator: 
\begin{equation}\label{eq:icl-ilr}
w_\theta: \mathbb{R}^{n(d+1)} \to \mathbb{R}^{d} \quad \text{such that} \quad (x_{1:n}, y_{1:n}) \mapsto \widehat{w} = w_\theta(x_{1:n}, y_{1:n}).
\end{equation}
This mapping $w_\theta$ is learned from a collection of $n_s$ distinct contexts, 
\begin{equation}\label{eqn:linearmodel}
\mathcal{D} = \{(x_{1:n}^j,y_{1:n}^j)\}_{j=1}^{n_s},  \quad \text{ where } y_i^j = \innerp{x_i^j,w^j}_{\R^d} 
    + \varepsilon_i^j, , \quad 1\leq j\le n_s.  
\end{equation}
These $w^j$ are drawn from a prior distribution $w^j \sim \mathcal{N}(w_0, \Sigma_w)$, and for each context, $x_i^j \sim \mathcal{N}(0, \Sigma_x)$ and $\varepsilon_i^j \sim \mathcal{N}(0, \sigma_\varepsilon^2)$. Crucially, while $n<d$, the weight vectors $w^j$ possess an intrinsic low-dimensionality, enforced by a low-rank prior covariance matrix $\Sigma_w$ with $\text{rank}(\Sigma_w) = r_w < n$.

To address this ill-posed inverse problem, the model must learn both a prior and an effective regularization strategy, and the learned map $w_\theta$ in \eqref{eq:icl-ilr} must be nonlinear in both $X=x_{1:n}$ and $Y=y_{1:n}$, since a properly regularized estimator depends nonlinearly on both. This is a key distinction from classical ridge regression approaches. A standard ridge estimator is typically tuned independently for each specific dataset $(X,Y)$ and does not leverage information from other related datasets. In contrast, the ICL-ILR map $w_\theta$ is a single function trained across multiple contexts. It learns to encode nonlocal information about the distribution of $w$ (i.e., the prior $\mathcal{N}(w_0, \Sigma_w)$) from the multi-task training data, and it uses this information for regularization. Table \ref{tab:paradigm_comparison} summarizes these three paradigms.


\textbf{Main results.}
\label{sec:main_results}
We demonstrate that a linear transformer, trained for the ICL-ILR task, successfully learns the inverse mapping $W(X,Y)$. Our primary contributions are threefold:

First, we show that the transformer implicitly learns both the prior and an effective regularization strategy from the training data. Specifically, our numerical results indicate that the transformer accurately captures the mean and the low-rank covariance of the prior. The learned regularization, emerging from this internalized prior, allows the transformer to significantly outperform a fine-tuned ridge regression (RE) and even a two-stage ridge estimator (TRE) that explicitly estimates the prior. 

Second, we establish that the intrinsic low-dimensionality of the tasks, characterized by the rank $r_w$ of the prior covariance $\Sigma_w$, is a critical factor for successful ICL in this regime. We numerically demonstrate that the estimation error deteriorates if the task dimension $r_w$ is not sufficiently smaller than the context length $n$. Even an oracle ridge estimator (ORE), which uses the true prior and noise distribution and is statistically optimal, struggles when $r_w \ge n$, underscoring that low task dimensionality is a fundamental requirement for resolving the underdetermined system.

Third, we numerically verify the scaling properties of the transformer's estimator. The error is shown to scale approximately linearly with the noise variance $\sigma_\varepsilon^2$, the ratio of the task dimension to the context length ($r_w/n$), and the condition number of the input data covariance $\Sigma_x$. These scaling laws align with the theoretically optimal ORE and further validate the transformer's ability to learn a robust solution to the ICL-ILR problem.

Collectively, these findings not only highlight the potential of transformers for solving ill-posed inverse problems but also offer insights into the mechanisms by which they extract and utilize statistical knowledge from contextual examples.

\subsection{Related work}
\label{sec:related_work}
Our research intersects three dynamic areas: the emergent field of ICL, ill-posed inverse problems with regularization, and the application of transformers to solve such problems by implicitly learning underlying statistical structures, often leveraging low-dimensional data representations. 

\textbf{ICL with Transformers.}
Transformers have demonstrated a remarkable ability for in-context learning in language models, adapting to new tasks from a few examples in the prompt without weight updates \cite{Brown2020Language}. Theoretical understanding of ICL is advancing rapidly. Several studies have demonstrated that transformers are versatile learners, capable of learning low-dimensional target functions, simple functions, linear models, and algorithms in context \cite{garg2022can,bai2023transformers,oko2024pretrained,zhang2024trained,vladymyrov2024linear,furuya2024transformers,bhattamishra2024understanding}. Also, ICL as gradient descent has been investigated in \cite{akyurek23learning,von2023transformers,mahankali24one,vladymyrov2024linear,ahn2023transformers}. Additionally, \cite{li2023transformers,collins2024context} investigates the generalizability and stability, and \cite{xie2022explanation, arora2024bayesian} suggest that transformers perform implicit Bayesian inference. In particular, ICL for linear regression has been extensively used in these studies,  with \cite{lu2024asymptotic} covering the asymptotic properties.  However, all these studies focus on the forward problem with $n>d$. In contrast, we focus on ICL for ILR in the rank-deficient scenarios with $n < d$, as reported to be challenging in \cite{ahuja2023transformers}. We show that transformers not only learn to mimic an algorithm but also infer the necessary prior and regularization strategy from the distribution of tasks.

\textbf{Regularization for Inverse Linear Regression.}
Rank-deficient ILR is an ill-posed problem requiring regularization to ensure a unique and stable solution. Classical methods include Tikhonov regularization, with ridge regression being a prominent example \cite{hoerl1970ridge}, where a penalty term stabilizes the estimate \cite{hansen1998rank,engl1996regularization}. The regularization parameter is typically chosen via data-dependent methods such as Generalized Cross-Validation (GCV) \cite{golub1979_GCV} or L-curve \cite{hansen00_Lcurve}. From a Bayesian perspective, such regularization often corresponds to imposing a prior distribution. These methods typically require explicit choices of regularization form and focus on estimates for each individual dataset. In contrast, ICL of ILR uses multiple training tasks to learn the inverse mapping. The two-stage ridge estimator benchmark in our study attempts to bridge this gap by first estimating a prior from the data and then applying classical regularization. The transformer, however, aims to learn both the prior characteristics and an effective regularization strategy in an end-to-end manner.

\textbf{Transformers for Inverse Problems.}
Transformers are increasingly employed to tackle diverse scientific inverse problems. Their strength lies in modeling complex dependencies within sequential or structured data, leading to novel applications in various domains \cite{guo2023Transformer, ovadia2024vito, yu2024nonlocal, evangelista2023ambiguity, chen2023deformable, cao2021choose,youYu2022_Datadriven}. These transformer-based approaches complement other deep learning strategies for inverse problems, such as those focused on learning regularization hyperparameters \cite{Afkham2021learning} or designing networks explicitly for stable inversion \cite{evangelista2025tobe}. This study, in particular, highlights that transformers possess the capability to learn implicit priors and regularization strategies directly from data, providing a promising foundation for future developments in solving ill-conditioned inverse problems.


\section{Linear transformer and benchmarks}
\label{sec:methods}
We first introduce the architecture of our proposed linear transformer, specifically designed to learn the ICL-ILR mapping. To evaluate its performance and understand its learning mechanism, we consider three benchmarks: (i) a standard ridge regression estimator (RE), representing a classical approach to regularized inverse problems; (ii) a two-stage ridge regression estimator (TRE), which attempts to extract prior information from the training data and utilize it for regularization; and (iii) an oracle ridge estimator (ORE). The ORE, which leverages the true prior and noise distributions, serves as a theoretical benchmark and is instrumental in highlighting the fundamental importance of low task dimensionality for effective estimation. Collectively, these benchmark estimators provide a context for dissecting and appreciating the capabilities of the linear transformer in learning the prior and regularization for the ICL-ILR task.

\subsection{Linear transformer for inverse linear regression}
\label{sec:Algo}
We define a \emph{linear transformer} as an L-layer neural network in which the first $(L-1)$ layers gather context information and the last layer achieves inversion with regularization. The first $(L-1)$ layers are linear self-attention layers responsible for processing the input context $E=(X,Y)$ and producing refined representation $E^{(L-1)}=(X^{(L-1)},Y^{(L-1)})$. The final $L$-th layer, which we term the \emph{inverse-regression layer}, then transforms these processed context tokens $E^{(L-1)}$ into the weight vector estimate $w_\theta(X,Y)$ with a linear attention mechanism. The last layer is crucial as it performs the actual inverse mapping for the inverse linear regression. 

Let $X=  (x_1,x_2,\ldots, x_n)^\top \in \R^{n\times d}$, $Y= (y_1,y_2,\ldots, y_n)^\top  \in \R^{n\times 1}$, and denote the token by  
$$E^{(0)}= (X,Y)
\in \R^{n\times (d+1)}. 
$$

Each of the $(L-1)$ linear self-attention layers consists of multiple heads, followed by a layer normalization. Each head is parameterized by three weight matrices and two bias vectors, namely, key $W_K\in\R^{d_k\times n}$, query $W_Q\in\R^{d_k\times n}$, value $W_V\in\R^{n\times n}$, and bias vectors $B_K, B_Q\in\R^{d_k\times 1}$. With the $l$-th layer being $E^{(l)}$, the update in the $(l+1)$-th layer with $H$ heads is  
\begin{equation}\label{eq:1head_update}
\textstyle{(\Delta E)^{(l)}} :=  \sum_{h=1}^H W_V^{(l,h)} E^{(l)}\big(W^{(l,h)}_K E^{(l)}+B^{(l,h)}_K\mathbf{1}_{d+1}\big)^\top \big( W^{(l,h)}_Q E^{(l)}+B^{(l,h)}_Q\mathbf{1}_{d+1}\big), 
\end{equation} 
where $\mathbf{1}_{d+1}=[1,\cdots,1]\in\R^{1\times (d+1)}$. 
 Then, a layer normalization applies to each row, which yields trainable parameters $\gamma^{(l)},\beta^{(l)}\in \mathbb{R} ^{(d+1)\times 1}$. The output of the $(l+1)$-th layer is 
\begin{equation}\label{eq:layers}
\begin{aligned}
   E^{(l+1)}=  &E^{(l)} +\text{layernorm}\left[(\Delta E)^{(l)}\right], \\ 
   \text{layernorm}\left[(\Delta E_i)^{(l)}\right]:=& \dfrac{1}{\widetilde{(\Delta E_i)^{(l)}}} \text{diag}(\gamma^{(l)}) \left((\Delta E_i)^{(l)}-\overline{(\Delta E_i)^{(l)}}\right)+\beta^{(l)}.
\end{aligned}
\end{equation}
Here, for a vector $a\in \R^{d+1}$, $\widetilde{a}$ and $\overline{a}$ are the standard deviation and mean of all elements in $a$. The parameter in the $l$-th layer is $\theta^{(l)} = \{W_Q^{(l,h)}, W_K^{(l,h)},W_V^{(l,h)},B_Q^{(l,h)}, B_K^{(l,h)}, \gamma^{(l)},\beta^{(l)}\}_{h=1}^H$. 


The last layer achieves inverse linear regression by transforming the context tokens $E^{(L-1)}$ into a $w$ estimator, hence providing an inverse mapping from the context $(X,Y)$ to the weight vector $w$:  
\begin{equation}\label{eq:inverse-layer}
	w_\theta(X,Y) = \big(W^{(L)}_K E^{(L-1)}+B^{(L)}_K\mathbf{1}_{d+1}\big)^\top \big( W^{(L)}_Q E^{(L-1)}+B^{(L)}_Q\mathbf{1}_{d+1}\big) \begin{pmatrix} W_{P,X} \\ W_{P,Y}  \end{pmatrix}. 
\end{equation} 
The parameter in this layer is $\theta^{(L)} = \{W_Q^{(L)}, W_K^{(L)},B_Q^{(L)}, B_K^{(L)}, W_{P,X}, W_{P,Y}\}$, where $W_{P,X}\in\R^{d\times 1}$ and $W_{P,Y}\in\R$.

\textbf{Training Process.}
The linear transformer model, with parameters $\theta = \{\theta^{(l)}\}_{l=1}^L$, is trained to learn the inverse mapping $w_\theta(X,Y)$ from the training datasets $\mathcal{D} = \{(X^j,Y^j)\}_{j=1}^{n_s}$. The training objective is to minimize the empirical mean-squared error of the reconstructed outputs:
\[ \mathcal{L}(\theta) = \frac{1}{n_s}\sum_{j=1}^{n_s}\| X^j w_\theta(X^j,Y^j) - Y^j\|_{\R^{n}}^2. \]
Herein, the value matrices in the self-attention layers are set to identity matrices ($W_V^{(l,h)}=I$) following \cite{yu2024nonlocal}. Key hyperparameters, including the key/query dimension $d_k$ and the number of layers $L$ are selected based on validation performance. Further details on training procedure strategies are provided in Appendix \ref{sec:additional_num}.

\textbf{Nonlinearity of the Inverse Mapping.}
Although termed a "linear transformer" due to its activation function, the learned inverse mapping $w_\theta(X,Y)$ is a highly nonlinear function of the input context $(X,Y)$. This nonlinearity arises from several sources: (i) the layer normalization applied after each attention update, (ii) the multi-layer composition of these operations, and (iii) the nonlinear final inverse-regression layer. 

This inherent nonlinearity is critical for the transformer's success in solving the rank-deficient ILR problem, as effective regularization inherently requires a nonlinear dependence on both $X$ and $Y$. The transformer's parameters, shaped by the training process, implicitly capture both the prior distribution of the weight vectors and this effective regularization strategy. To empirically demonstrate these learned capabilities and to understand the nature of the learned solution, we compare the performance of our linear transformer against two practical ridge regression estimators and the statistically optimal Bayesian estimator (Oracle Ridge Estimator), which are introduced in subsequent sections.

\subsection{Benchmarks: ridge estimators}
We consider two benchmarks: the classical regression estimator (RE) and a two-stage ridge estimator (TRE).  The RE applies ridge regression independently to each context, without leveraging any information from other tasks. In contrast, the TRE first infers a Gaussian prior over the weight vector from the training dataset and then uses this learned prior to regularize each context‐specific estimator.

Because $n<d$, the RE’s performance, even with optimally tuned hyperparameters, remains constrained by the rank deficiency of the inverse problem. The TRE, on the other hand, is asymptotically optimal as the number of training contexts grows, since it exploits the full prior information embedded in the training data.

\textbf{Benchmark 1: Ridge Estimator (RE).} The ridge regression estimator for each context $(X,Y)=(x_{1:n},y_{1:n})$ is    
\[
\widehat{w}_\lambda= ( X^{\top} X  + n \lambda I )^{-1} X^{\top}  Y,  
\]
which comes from the model $ Y = Xw + \varepsilon$. Here, the hyperparameter $\lambda$ is fine-tuned for each dataset, selected by the the GCV method of \cite{golub1979_GCV} as the minimizer of 
$
GCV(\lambda) = \frac{\|Y-\widehat Y_\lambda\|^2}{\left( n - \mathrm{Trace}(A(\lambda)) \right)^2  }= \frac{\| ( I - A(\lambda) ) Y\|_{\R^d}^2/n}{ \left( n - \mathrm{Trace}(A(\lambda)) \right)^2  }$, where $A(\lambda) = X  ( X^\top  X + n \lambda I )^{-1} X^\top $.  

\textbf{Benchmark 2: Two-stage Ridge Estimator (TRE).}  The two-stage ridge regression estimator first infers a Gaussian prior for the weight vector from the training dataset and then uses this prior as the regularizer of the ridge regression. 

In TRE, the first stage extracts the mean and covariance of a Gaussian prior $\mathcal{N}(\widehat{w}_0,\widehat \Sigma_w)$ from the empirical mean and covariance of the least squares estimators (LSE) with a minimal norm for the training contexts. In particular, since the LSE's covariance matrix is $\Sigma_w + \sigma_\varepsilon^2 \E[(X^\top X)^\dag]$ with an unknown $\sigma_\varepsilon^2$, we use eigen-decomposition of the LSE's empirical covariance matrix to estimate $\widehat \Sigma_w$; see Section \ref{append:TRE}.     

In the second stage, one uses the estimated prior $\mathcal{N}(\widehat{w}_0,\widehat \Sigma_w)$ for ridge regression with $\lambda$ tuned for each test context $(X,Y)$. Specifically, using the learned prior, the model can be written as $Y= X(\widehat{w_0}+ \widehat \Sigma_w^{1/2} v) + \varepsilon$ with $v\sim \mathcal{N}(0, I_d)$. Then, its ridge estimator is    
\begin{equation}\label{eq:tRE}
\begin{aligned}
	\widetilde{w}^{tRE}_\lambda  = \widehat{w_0}+ \widehat{\Sigma}_w^{1/2} \widehat{v}_\lambda,  \quad 
	\widehat{v}_\lambda & = \argmin{v\in \R^d } \|X \widehat{\Sigma}_w^{1/2} v- (Y-X \widehat{w_0})\|^2  + \lambda v ^\top v
	\\ & =   \left(\widehat{\Sigma}_w ^{1/2}X^\top X \widehat{\Sigma}_w ^{1/2} + n\lambda I_d \right)^{-1} (\widehat{\Sigma}_w^{1/2} X^\top (Y- X\widehat{w_0}) ). 
	 \end{aligned}
\end{equation} 
Here, the hyperparameter $\lambda$ is selected by the GCV method. In practice, one uses the eigenvalues and eigenvectors to compute $\widehat \Sigma_w^{1/2} = \sum_{k=1}^{\widehat r_w} ( \widehat \lambda_k^w)^{1/2}  v_k v_k^\top$, which enhances the stability of the TRE.

\subsection{Oracle ridge estimator and error bounds}
\label{sec:ORE}
To establish a theoretical performance benchmark, we define an Oracle Ridge Estimator (ORE). This estimator represents the optimal performance achievable under our problem settings, as it utilizes the true prior distribution $\mathcal{N}(w_0, \Sigma_w)$ and the true noise distribution with $\mathcal{N}(w_0, \sigma_\varepsilon^2 I_n)$, which are typically unknown in practice,  to obtain a regularized estimator for each context. Formally, the ORE is the posterior mean of the weight vector given the prior of $w$ and the likelihood of $(X,Y)$.  

Specifically, let $X\in \R^{n\times d}$ have rows being independent $\mathcal{N}(0,\Sigma_x)$ distributed, and let 
\begin{equation}\label{eq:model_array}
  Y = Xw + \varepsilon,
  \quad w \sim \mathcal{N}(0,\Sigma_w),
  \quad \varepsilon \sim \mathcal{N}(0,\sigma_\varepsilon^2 I_n),
\end{equation}
where $\varepsilon$ and $w$ are independent, and $\Sigma_w = U\,\Lambda\,U^\top$ with $U\in\R^{d\times r_w}$ being column orthonormal and $\Lambda \in \R^{r_w\times r_w} 
$ being a diagonal matrix of positive eigenvalues. 

The ORE is the posterior mean of the weight vector given $(X,Y)$. We denote it by $\widehat{w}^{\mathrm{ORE}}= \widehat{w}^{\mathrm{ORE}}(X,Y)= \mathbb{E}[w|X,Y]$. A direct computation shows that it has the representation    
\begin{equation}\label{eq:ORE}
  \widehat{w}^{\mathrm{ORE}} = U \widehat v, \quad 
  \widehat v =  \Sigma^{\mathrm{post} }_v \tfrac1{\sigma_\varepsilon^2}(XU)^\top Y, \quad \Sigma^{\mathrm{post} }_v = \big( \Lambda^{-1} +\tfrac1{\sigma_\varepsilon^2}(XU)^\top XU\big)^{-1}.  
\end{equation}
Here, the posterior of $w$ has a degenerate covariance, so we consider the projected vector $v$ in $\R^{r_w\times 1}$, whose posterior is $\mathcal{N}(\widehat v, \Sigma^{\mathrm{post} }_v)$ with $\Sigma^{\mathrm{post} }_v\in \R^{r_w\times r_w}$ being strictly positive definite.

The next two propositions present error bounds for the conditional and unconditional Bayes risks of the ORE, which scale at the order $O(\frac{\sigma_\varepsilon^2\,r_w}{n})$. We postpone their proofs to Section \ref{append:proofs}. Here, we use the Lownner order $A\succeq B\succeq 0$ to denote that $x^\top (A - B)x \geq 0$ for any $x$ when $A,B\in \R^{d\times d}$ are positive semi-definite (PSD) matrices. We use $\mathbb{E}_{w,\varepsilon}$ to denote the expectation with respect to $w,\varepsilon$. 

\begin{proposition}[Conditional Bayes risk of ORE]
\label{prop:bd_ore_condX}
Conditional on a given $X\in \R^{n\times d}$ for the model in \eqref{eq:model_array},  
 the Oracle Ridge Estimator in \eqref{eq:ORE} has Bayes risk bounded by 
$  \mathbb{E}_{w,\varepsilon}\bigl\|\widehat{w}^{\mathrm{ORE}} - w\bigr\|^2
  = \mathbb{E}_{w,\varepsilon}\bigl\|\widehat v -  v\bigr\|^2
  = \mathrm{Tr}\bigl(\Sigma_v^{\mathrm{post} }\bigr).  
$ 
In particular, when $
\tfrac1n U^\top X^\top XU\succeq  b I_{r_w}$ with $b>0$, we have 
\begin{equation}\label{eq:ore_risk_bound1}
    \mathbb{E}_{w,\varepsilon}\bigl\|\widehat{w}^{\mathrm{ORE}} - w\bigr\|^2
  \;\le\; \frac{\sigma_\varepsilon^2\,r_w}{n\,b} . 
\end{equation}
\end{proposition}

\begin{proposition}[Bounds for Bayes risk of ORE]
\label{prop:2sideBd_ore}
For any $0\le t < 1-\sqrt{r_w/n}$, the Oracle Ridge Estimator in \eqref{eq:ORE} for Model \eqref{eq:model_array} 
 has Bayes risk bounded by 
\begin{equation}\label{eq:ore_risk_bound_2sides}
\begin{aligned}
\frac{r_w\,\sigma_\varepsilon^2\,\lambda_{\min}(\Sigma_w)c_{t}}
     {\sigma_\varepsilon^2 + n\,\lambda_{\min}(\Sigma_w)\,\lambda_{\max}(\Sigma_x)\,a_t} \leq   \mathbb{E}\bigl\|\widehat{w}^{\mathrm{ORE}} - w\bigr\|^2 
   \le  \frac{\sigma_\varepsilon^2\,r_w}{n\,\lambda_{\min}(\Sigma_x)b_t} \;+\;
r_w\,\lambda_{\max}(\Sigma_w) 2\,e^{-n t ^2/2}. 
\end{aligned}
\end{equation}
where $c_{t}= 1-2e^{-nt^2/2} $, $a_t= (1+\sqrt{r_w/n}+t )^2$ and $b_t= (1-\sqrt{r_w/n}-t )^2$. 
\end{proposition}

 \begin{remark}[Asymptotics of the ORE in context length] When $n$ is large and $r_w/n <1/2$, the Bayes risk of the ORE scales like $\frac{ r_w\,\sigma_\varepsilon^2}{n} $, i.e., 
 \begin{equation*}
 \mathbb{E}\bigl\|\widehat{w}^{\mathrm{ORE}} - w\bigr\|^2 =  O\bigl(\frac{ r_w\,\sigma_\varepsilon^2}{n\,\kappa}\bigr), 
 \end{equation*}
  where $\kappa=\mathrm{cond}(\Sigma_x)$ is the condition number of $\Sigma_x$.  
 In fact, choosing $t =2\sqrt{\ln n / n}$ gives  
$ e^{-nt^2/2}=O(n^{-2})$, $a_t = (1+\sqrt{r_w/n}+t)^2 < 4$, $b_t = (1-\sqrt{r_w/n}-t)^2 > 1/4$. Thus, for large $n$, Eq.\eqref{eq:ore_risk_bound_2sides} implies  
$\mathbb{E}\bigl\|\widehat{w}^{\mathrm{ORE}} - w\bigr\|^2 
\gtrsim  \frac{r_w\,\sigma_\varepsilon^2\,}
     {\sigma_\varepsilon^2 \lambda_{\min}(\Sigma_w)^{-1}+ 4 n\,\lambda_{\max}(\Sigma_x)}
\;\Bigl(1 - O(n^{-2})\Bigr) =  \frac{r_w\,\sigma_\varepsilon^2}{n\,\lambda_{\max}(\Sigma_x)} O(1) $ and $\mathbb{E}\bigl\|\widehat{w}^{\mathrm{ORE}} - w\bigr\|^2 
\lesssim  \frac{ \sigma_\varepsilon^2\,r_w}{n\,\lambda_{\min}(\Sigma_x) b_t} \;+\;
 O(n^{-2})
\lesssim 
\frac{4 r_w\,\sigma_\varepsilon^2}{n\,\lambda_{\min}(\Sigma_x)}.$
These two bounds show that the Bayes risk scales like
$O\bigl(\frac{ r_w\,\sigma_\varepsilon^2}{n\,\kappa}\bigr)$.
 \end{remark}

 The next lemma shows that the mean of the ORE is the same as the prior mean, and the covariance of ORE is close to the prior covariance when the noise variance is small.  
 \begin{lemma}[Mean and Covariance of the ORE]
 \label{lemma:cov_ORE}
The Oracle Ridge Estimator $\widehat{w}^{\mathrm{ORE}}(X,Y)$ in \eqref{eq:ORE} for Model \eqref{eq:model_array} has mean $\E[\widehat{w}^{\mathrm{ORE}}]= w_0$ and covariance 
$$ \text{Cov}(\widehat{w}^{\mathrm{ORE}}) = \Sigma_w - U \mathbb{E}_{X}[\Sigma_v^{\mathrm{post}}]  U^\top = \Sigma_w -\sigma^2_\varepsilon C, 
$$
where $\Sigma_v^{\mathrm{post}} $ is defined in \eqref{eq:ORE} and $C=U \mathbb{E}_{X}\left[ \bigl( \sigma^2_\varepsilon \Lambda^{-1} + U^\top X^\top X U \bigr)^{-1}\right] U^\top$. 
\end{lemma}

\section{Transformer learns the prior and regularization} \label{sec:prior_regu}
The linear transformer, as defined in Section \ref{sec:methods}, is trained to learn the inverse mapping $w_\theta(X,Y)$. A central hypothesis of this work is that for the rank-deficient ILR ($n<d$), the transformer implicitly recovers both the prior distribution $\mathcal{N}(w_0, \Sigma_w)$ from which the tasks $w^j$ are drawn, and an effective regularization strategy that leverages this learned prior. This section presents experimental results to validate these capabilities. In all experiments of this section, we fix $d=100$, $n=50$, and $\sigma_w=0.01$.

\subsection{Learning prior distribution} \label{sec:prior}

\begin{figure}[h!]
    \centering
   \includegraphics[width=0.98\textwidth]{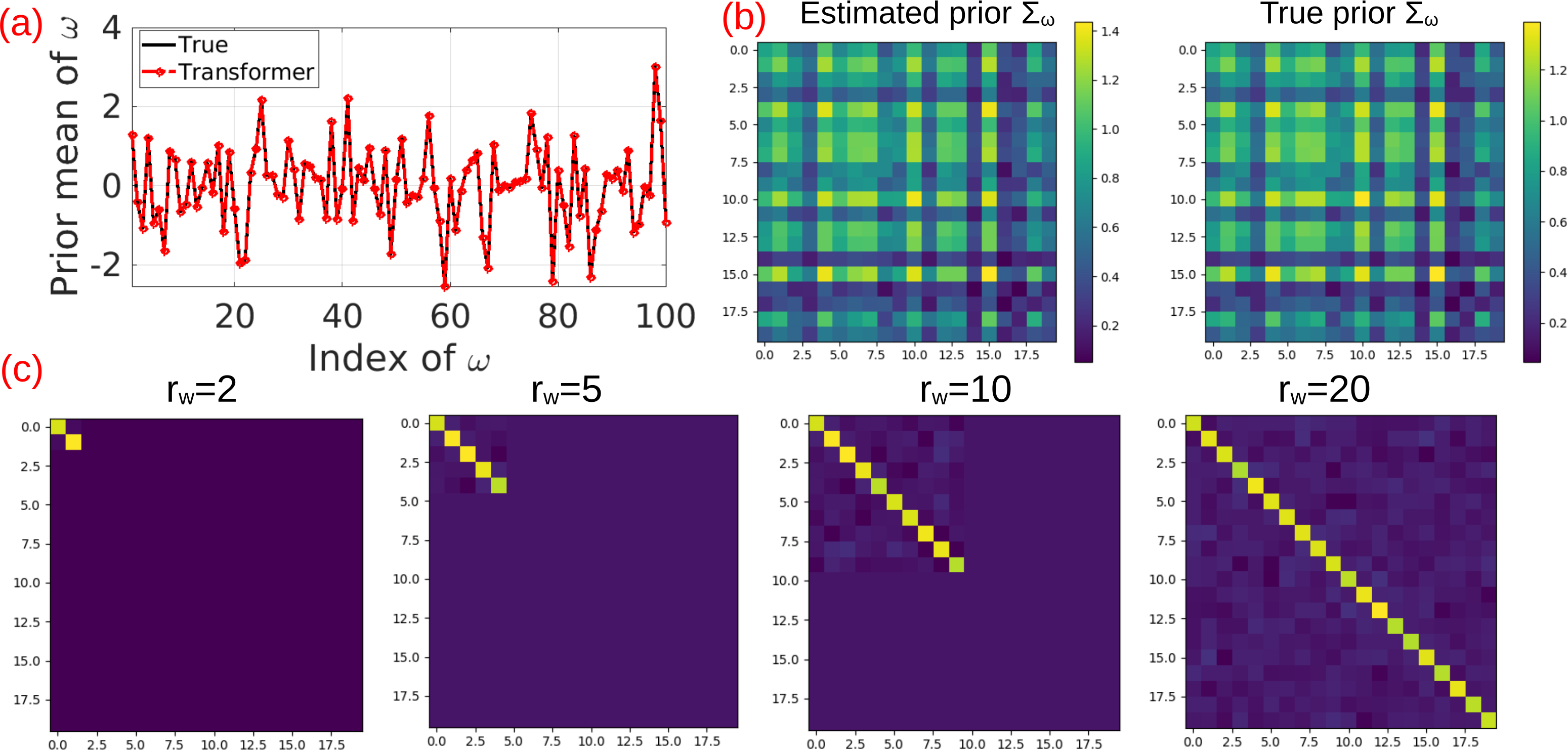}
    \caption{Prior mean and covariance matrices implicitly learned by Transformer. (a): Prior mean. (b): Learned non-diagonal prior covariance $\Sigma_w$. (c) Learned diagonal $\Sigma_w$ with varying rank $r_w$. They demonstrate the transformer's ability to recover the hidden prior and task dimensions.}
    \label{fig:learnprior}
\end{figure}

\begin{figure}[h!]
    \centering
    \includegraphics[width=0.66\textwidth]{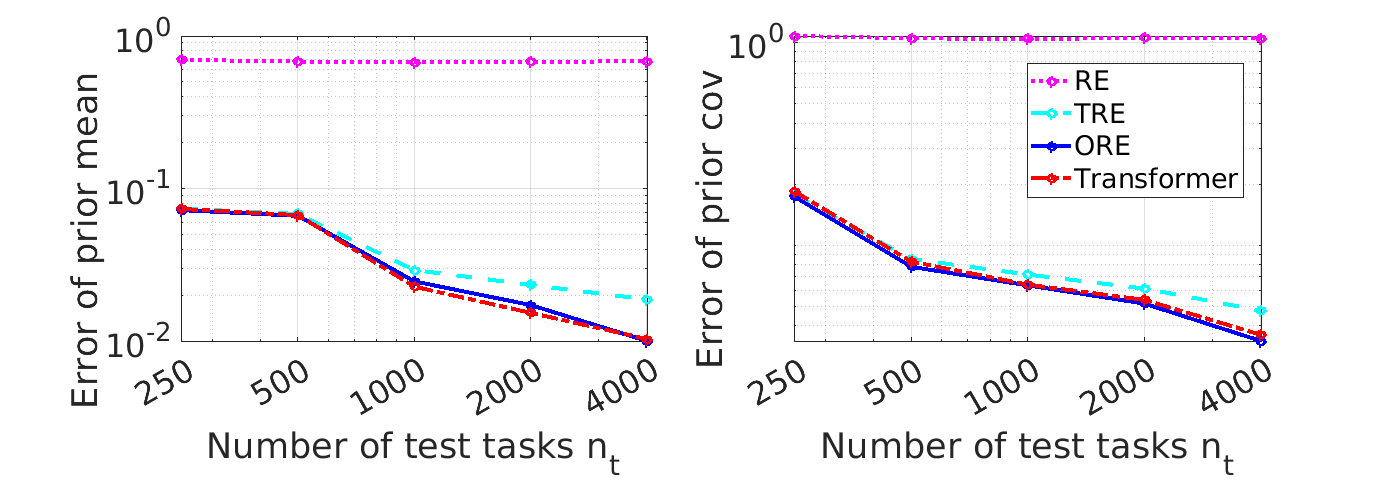}
    \includegraphics[width=0.32\textwidth]{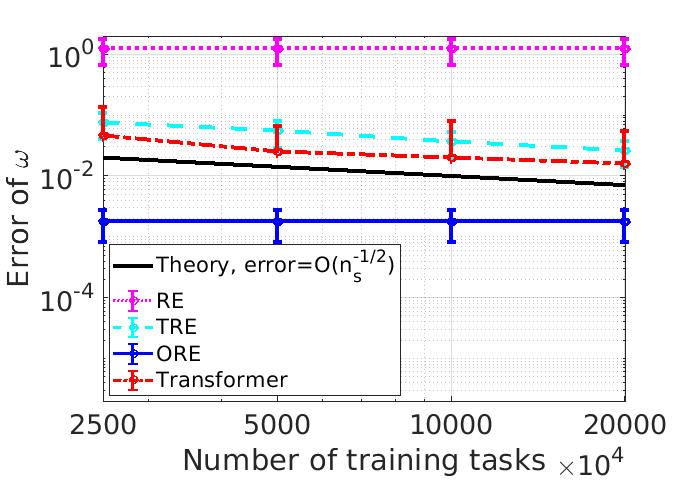}
    \caption{Error decay in the number of tasks in testing and training. Left and Middle: errors of estimator for prior mean and covariance. Transformer outperforms RE and TRE, matching ORE. Right: errors of $w$ estimators. Transformer outperforms TRE due to implicit regularization. 
    }
    \label{fig:prior_error}
\end{figure}

We first verify that the transformer $w_\theta(X,Y)$ internalizes the Gaussian prior $\mathcal{N}(w_0,\Sigma_w)$.  Since the trained transformer $w_\theta(X,Y)$ closely matches the oracle posterior mean $\E[w\mid X,Y]$ (the ORE), whose population mean and covariance are $
\E[\E[w\mid X,Y]] = w_0$ and $
\mathrm{Cov}\bigl(\E[w\mid X,Y]\bigr) = \Sigma_w - \sigma_\varepsilon^2\,C
$
as in Lemma \ref{lemma:cov_ORE}, we can recover the prior parameters by the statistics of the transformer outputs.  
In particular,  $w_0\approx \E[w_\theta(X,Y)]$ and  
\[
 \Sigma_w \approx   \text{Cov}({w}_\theta(X,Y))  +\sigma^2_\varepsilon C, \text{ where }C= U \mathbb{E}_{X}\left[ \bigl( \sigma^2_\varepsilon \Lambda^{-1} + U^\top X^\top X U \bigr)^{-1}\right] U^\top.
\]
Thus, on a set of new and unseen test contexts $\{(\tilde{X}^j, \tilde{Y}^j)\}_{j=1}^{n_t}$, 
we compute the empirical mean and covariance of the transformer estimates, 
\begin{align*}
\bar{w}_\theta  = \frac{1}{n_{t}}\sum_{j=1}^{n_{t}} w_\theta(\tilde{X}^j, \tilde{Y}^j), \quad 
\widehat{\Sigma}_{w_\theta}  = \frac{1}{n_{t}-1}\sum_{j=1}^{n_{t}} (w_\theta(\tilde{X}^j,\tilde{Y}^j) - \bar{w}_\theta)(w_\theta(\tilde{X}^j,\tilde{Y}^j) - \bar{w}_\theta)^\top.
\end{align*}
Then, we compare $\bar{w}_\theta $ and the estimated prior covariance matrix $ \widehat{\Sigma}_{w_\theta} + \widehat C $ with with the ground-truth prior mean $w_0$ and covariance $\Sigma_w$. Here, $\widehat C$ is an empirical approximation of the matrix $\sigma^2_\varepsilon C$.

In Figure \ref{fig:learnprior}, both training and test tasks use
$x_i^j\sim\mathcal{N}(0,I_d)$, $w^j\sim\mathcal{N}(w_0,\,U U^\top),
$ with $U\in\R^{d\times r_w}$ having orthonormal columns, and outputs $y_i^j=\langle x_i^j,w^j\rangle+\varepsilon_i^j$ per \eqref{eqn:linearmodel}.  Panel (a) shows that $\bar w_\theta$ nearly coincides with $w_0$.  Panel (b) plots $\widehat\Sigma_{w_\theta}+\widehat C$ against $\Sigma_w$, demonstrating excellent agreement. Finally, in panel (c) we vary the intrinsic rank $r_w\in\{2,5,10,20\}$ with $\Sigma_w=I_{r_w}$, and observe that $\widehat\Sigma_{w_\theta}+\widehat C$ exhibits exactly $r_w$ dominant eigenvalues before a sharp spectral drop, showing that the transformer correctly recovers the true task dimension and structure. 

To quantify the approximation, the two left panels of Figure \ref{fig:prior_error} plot the estimation error of the prior mean and covariance versus the number of test contexts $n_t$.  The transformer’s estimates converge significantly faster and more accurately than those of the RE and TRE baselines, matching the estimates produced by the ORE.

These results strongly suggest that the transformer implicitly learns the mean and covariance of the prior distribution. This acquired knowledge about the distribution of tasks is foundational for its ability to effectively regularize the ill-posed inverse problem.

\subsection{Transformer learns the regularization}\label{sec:regu}

We show next that the transformer's learned inverse mapping $w_\theta(X,Y)$ inherently regularizes the problem, leading to improved performance as the training task size increases. In the right plot of Figure \ref{fig:prior_error}, we report the mean and standard deviation of the error $\| \widehat w (X,Y) - w \|$
over $n_t=1000$ unseen test contexts. Here, we consider $r_w=2$ and increased training context size $n_s=2500\sim 20000$.
One can see that the overall performance ranking is consistently $Err(RE) \gg Err(TRE) \gtrsim Err(w_\theta) \gtrsim Err(ORE)$.

In particular, because the RE is tuned independently in each context without leveraging the shared prior structure across tasks, its performance is flat in $n_s$ and significantly behind the other estimators. On the other hand, the TRE improves with more data, highlighting the advantage of explicitly estimating the prior in the first stage, but it still lags behind the transformer, which jointly learns the prior and regularization end-to-end.  As $n_s$ grows, the transformer’s RMSE approaches the oracle ridge estimator’s, showing that its learned regularization nearly matches the optimal performance achievable with the true prior.

\textbf{Nature of the Learned Regularization.} 
The transformer's ability to regularize effectively stems from its architecture, particularly the non-linear operations involving attention mechanisms and layer normalization. The training process, which minimizes the prediction error over numerous contexts drawn from the same prior, forces the parameters $\theta$ to encode this prior information. The resulting mapping $w_\theta(X,Y)$ is a complex non-linear function of the input context $(X,Y)$. This allows for a more nuanced regularization than a simple quadratic penalty (as in ridge regression). The attention mechanism can adaptively weigh different parts of the context $(X, Y)$, and the multi-layer structure enables hierarchical feature extraction, all of which contribute to a data-dependent regularization that is implicitly tuned to the learned prior. While the RE and TRE rely on tuning a single parameter $\lambda$, the transformer's regularization is an intrinsic property of the learned mapping $w_\theta$.

\section{Low-dimensionality of tasks and scaling}
\label{sec:lowD-scaling}

\begin{figure}[h!]
    \centering
    \includegraphics[width=0.98\textwidth]{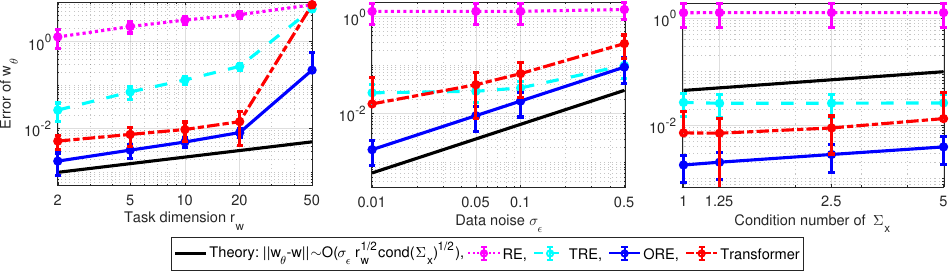}
    \caption{Scaling of the error in task dimension $r_w$, data noise level $\sigma_\varepsilon$, and condition number of input data covariance matrix $\Sigma_x$. The transformer achieves scalings aligning with ORE. }
    \label{fig:scaling}    
\end{figure}

In this section, we further numerically investigate the dependence of the linear transformer's performance on this task dimensionality $r_w$ and other fundamental problem parameters, including the noise level $\sigma_\varepsilon^2$ and the conditioning of the input data $X$. The theoretical error bounds for the ORE in Proposition \ref{prop:2sideBd_ore} suggest that the optimal Bayes risk scales with the ratio $r_w/n$, and is influenced by $\sigma_\varepsilon^2$ and $\kappa(\Sigma_x)$. We aim to numerically verify if the learned linear transformer exhibits similar scaling behaviors, which would further underscore its ability to learn an effective inverse mapping. In all experiments of this section, we again fix $d=100$ and $n=50$.

\textbf{Low task dimensionality.}
We first investigate the impact of task dimensionality, by varying the rank $r_w$ of the prior covariance matrix $\Sigma_w= \mathrm{Diag}(I_{r_w}, 0, \ldots, 0)$ from $2$ to $50$, and keeping $\sigma_\varepsilon=0.01$ and $\Sigma_x=I_d$ fixed. Then, we evaluate the performance of each model over $1000$ test contexts, and report the mean and standard deviation of $\| \widehat w (X,Y) - w \|$ in the left plot of Figure \ref{fig:scaling}. The estimation error of all methods grows as $r_w$ increases relative to $n$. This confirms that accurate recovery hinges on having a low task dimension ($r_w < n$).
Notably, the transformer outperforms even the TRE, indicating its ability to harness low‐dimensional structures more effectively.  Moreover, the error scales roughly like $\sqrt{r_w}$, in line with the $O(r_w/n)$ dependence predicted by Proposition \ref{prop:2sideBd_ore}. 
Furthermore, once $r_w$ approaches or exceeds $n=50$, all estimators, including the ORE, suffer a steep rise in error.  Even with the true prior in hand, $n$ samples cannot resolve $r_w$ degrees of freedom when $r_w\ge n$, making the inverse problem fundamentally under-determined.  This behavior confirms the necessity of $r_w<n$ for successful ICL, as argued in Section \ref{sec:ORE}. 

These results underscore that the low dimensionality of tasks ($r_w < n$) is a fundamental prerequisite for the transformer to learn effectively from context in rank-deficient ILR. 

\textbf{Scaling with Noise Level ($\sigma_\varepsilon$).}
The middle panel of Figure \ref{fig:scaling} plots the estimation error against the noise standard deviation $\sigma_\varepsilon$, holding $r_w=2$ and $\Sigma_x=I_d$. Across the transformer, ORE, and TRE, the error grows approximately linearly in $\sigma_\varepsilon$, matching the ORE’s error bound.  This confirms that the transformer’s learned mapping exhibits the correct sensitivity to observation noise.

\textbf{Scaling with Condition Number ($\kappa(\Sigma_x)$):}
The right panel of Figure \ref{fig:scaling} shows error versus the condition number $\kappa(\Sigma_x)$, with $r_w=2$ and $\sigma_\varepsilon=0.01$.  All three methods exhibit increasing error as $\kappa(\Sigma_x)$ grows. This matches the ORE's theoretical bounds, which worsen as $\lambda_{\min}(\Sigma_x)$ decreases (and hence $\kappa(\Sigma_x)$ increases). Although the exact dependence may be intricate, the clear trend confirms that better‐conditioned inputs improve performance and that the transformer's learned mapping correctly incorporates this sensitivity into its implicit regularization.

\section{Conclusion}
 We have demonstrated that even a simple linear transformer can, through in-context learning, internalize the cross-task prior and optimal regularization needed to solve rank-deficient inverse linear regression.  By matching oracle error bounds and outperforming classical per-task ridge methods, the transformer showcases its ability to extract and exploit low-dimensional structure, adaptively balancing noise sensitivity and data conditioning.

\textbf{Limitations.} Our analysis has focused on the Gaussian ILR setting and a linear transformer; extending these findings to nonlinear transformers and non-Gaussian priors remains an open challenge. The current framework also assumes access to large numbers of training tasks for meta-learning the prior, which may be restrictive in domains with scarce data. Additionally, the ill-posedness in this study stems from a rank deficiency, leaving it open to investigation of ICL for ill-conditioned inverse problems. Finally, while our experiments confirm the correct scaling laws, a tighter theoretical bound for transformers is still open.

{\bf Broader Impacts.}  Nevertheless, these findings illuminate a powerful new paradigm: transformers as meta-solvers for ill-posed inverse problems.  Rather than hand-crafting priors or tuning regularizers for each task, a single model can learn to perform Bayesian-style inference on the fly, opening pathways to tackle nonlinear dynamics, partial differential equations, and high-resolution imaging.  While our work focuses on Gaussian ILR and linear attention, it lays the groundwork for more complex architectures and broader domains.  In-context learning for inverse problems is still in its infancy, and we are only beginning to uncover its vast potential.  


\begin{ack}

Y.~Yu would like to acknowledge support by the National Science Foundation (NSF) under award DMS-2436624, the Department of the Air Force under award FA9550-22-1-0197, and the National Institute
of Health under award 1R01GM157589-01. Portions of this research were conducted on Lehigh University's Research Computing infrastructure partially supported by NSF Award 2019035. F.~Lu would like to acknowledge support by NSF CAREER DMS-2238486. 
\end{ack}



{\small
\bibliographystyle{plain}

}

\newpage

\appendix

\section{Technical Appendices and Supplementary Material}

\subsection{The two-stage ridge estimator}\label{append:TRE}
In the process of constructing the TRE, one first extracts the mean and covariance of a Gaussian prior $\mathcal{N}(\widehat{w}_0,\widehat \Sigma_w)$ from the empirical mean and covariance of the least squares estimators (LSE) with a minimal norm for the training contexts. 
That is, for each context $(X^j,Y^j)$ in the training dataset, the LSE is 
$$
\widehat{w}^j = ( X^{j,\top} X^j  )^{\dag} X^{j,\top} Y^{j}, 
\quad, 1 \leq j\leq n_s. 
$$ 
By the strong law of large numbers, their empirical mean and covariance approximate the true prior mean and covariance as the sample size $n_s$ increases (see Lemma \ref{lemma:TRE_mean_cov}), 
\begin{align*}
	\widehat{w}_{0} = \frac{1}{n_s}\sum_{j=1}^{n_s} \widehat{w}^j \to w_0; \quad 
	\widehat C =   \frac{1}{n_s-1}\sum_{j=1}^{n_s} ( \widehat{w}^j-\widehat{w_0} ) (\widehat{w}^{j} -\widehat{w_0} )^\top \to \Sigma_w + \sigma_\varepsilon^2 \E[(X^\top X)^\dag]. 
\end{align*}
Thus, an estimator for the prior covariance is $\widehat \Sigma_w = \widehat C-\sigma_\varepsilon^2 \widehat{\Sigma_x^{-1}}
$, where $\widehat{\Sigma_x^{-1}} = \frac{1}{n_s} \sum_{j=1}^{n_s} ( X^{j,\top} X^j )^{\dag} $. Since $\sigma_\varepsilon^2$ is unknown and it cannot be estimated from the residual of the LSEs due the rank deficiency, we estimate $\widehat \Sigma_w$ by eigen-decomposition of $\widehat C$ assuming that $\widehat{\Sigma_x^{-1}} $ is near $\mathrm{Diag}(I_n,0)$ to simply the computation. Denote eigenvalues of $\widehat C$ by $\lambda_k^C$ in descending order and denote $v_k$ the corresponding eigenvectors. When the sample size $n_s$ is large, the eigenvalues $\{\lambda_k^C\}$ will have two spectral gaps, one at $k=r_w$ and another at $k=n$, which can be detected with thresholds near $\lambda_{min}(\Sigma_w)$ and $\sigma_\varepsilon^2$, respectively. Let $\widehat \sigma_{\varepsilon}^2$ be the average of eigenvalues between the two gaps. Then, the eigenvalues of $\Sigma_w$ are estimated by $\widehat \lambda_k^w= \lambda_k^C - \widehat \sigma_{\varepsilon}^2$ for $1\leq k\leq  r_w$ and $\widehat \Sigma_w = \sum_{k=1}^{ r_w}\widehat \lambda_k^w v_k v_k^\top$.

The next lemma shows that the prior mean and covariance estimators in the two-stage ridge estimator converge to the true prior mean and covariance. 
\begin{lemma}\label{lemma:TRE_mean_cov}
Let $\{(X^j,Y^j)\}_{j=1}^{n_s}$ be samples of the model
\begin{equation*}
  Y = Xw + \varepsilon,
  \quad w \sim \mathcal{N}(w_0,\Sigma_w),
  \quad \varepsilon \sim \mathcal{N}(0,\sigma_\varepsilon^2 I_n),
\end{equation*}
where $\varepsilon$ and $w$ are independent, and $\Sigma_w$ has a rank $r_w<n$, and $X\in \R^{n\times d}$ has independent $\mathcal{N}(0,\Sigma_x)$ distributed rows. 
Let $\widehat{w}^j$ be the least squares estimator for each of the $(X^j,Y^j)$ pair:  
$$
\widehat{w}^j = ( X^{j,\top} X^j  )^{\dag} X^{j,\top} Y^{j}, 
 \quad 1 \leq j\leq n_s.   
$$ 
Then, as $n_s\to \infty$, the empirical mean and covariance of $\{\widehat{w}^j\}$ converge almost surely:  
\begin{align*}
	\widehat{w_0} &= \frac{1}{n_s}\sum_{j=1}^{n_s} \widehat{w}^j   \to  w_0,   \,\,  
		\\
	\widehat C&=   \frac{1}{n_s-1}\sum_{j=1}^{n_s} ( \widehat{w}^j-\widehat{w_0} ) (\widehat{w}^{j} -\widehat{w_0} )^\top \to   \Sigma_w + \sigma_\varepsilon^2 \E[( X^\top X)^\dag ].  
\end{align*}	
\end{lemma}
\begin{proof}
Note that 	
$$
\widehat{w}^j = ( X^{j,\top} X^j  )^{\dag} X^{j,\top} Y^{j} = w^j +  ( X^{j,\top} X^j  )^{\dag} X^{j,\top} {\varepsilon^j}, \quad 1 \leq j\leq n_s.   
$$ 
The samples $w^j$ are drawn from the distribution $\mathcal{N}(w_0,\Sigma_w)$, so by the strong law of large numbers, 
$$
\frac{1}{n_s}\sum_{j=1}^{n_s} w^j\to  w_0, \quad  
\frac{1}{n_s}\sum_{j=1}^{n_s-1} ( w^j- w_0)( w^j- w_0)^\top \to \Sigma_w, 
$$ 
almost surely as $n_s\to \infty$. Similarly, as $n_s\to \infty$, we have 
 \begin{align*}
 	& \frac{1}{n_s}\sum_{j=1}^{n_s} ( X^{j,\top} X^j  )^{\dag} X^{j,\top} {\varepsilon^j}\to \E[(X^\top X)^\top X^\top \varepsilon]=0 \\
 	& \frac{1}{n_s-1}\sum_{j=1}^{n_s} ( X^{j,\top} X^j  )^{\dag} X^{j,\top} {\varepsilon^j}(\varepsilon^j)^\top X^{j}( X^{j,\top} X^j  )^{\dag}\to \E[( X^\top X)^\dag ], 
 \end{align*}
where the second limit follows from $\E[ \varepsilon \epsilon^T\mid X]= I_n$ and  
\begin{align*}
   \E[( X^\top X)^\dag X^\top \varepsilon \epsilon^T X( X^\top X)^\dag ] 
=  \E[( X^\top X)^\dag X^\top \E[ \varepsilon \epsilon^T\mid X] X( X^\top X)^\dag ]
=  \E[( X^\top X)^\dag ]
\end{align*}
 due to the independence between $X$ and $\varepsilon$. 

Therefore, combining the above limits, and using the independence between $X,w$ and $\epsilon$, we obtain 
\begin{align*}
	\widehat{w_0} &= \frac{1}{n_s}\sum_{j=1}^{n_s} \widehat{w}^j =  \frac{1}{n_s}\sum_{j=1}^{n_s} [ w^j +  ( X^{j,\top} X^j  )^{\dag} X^{j,\top} \vec{\varepsilon^j} ]  \to w_0, \\
	\widehat C&=   \frac{1}{n_s-1}\sum_{j=1}^{n_s} ( \widehat{w}^j-\widehat{w_0} ) (\widehat{w}^{j} -\widehat{w_0} )^\top   \to  \Sigma_w + \sigma_\varepsilon^2  \E[( X^\top X)^\dag ] 
\end{align*}
as $n_s\to \infty$. 
\end{proof}

\subsection{Proofs for error bounds of ORE} \label{append:proofs}

Next, we prove the error bounds for the conditional and unconditional Bayes risk of the ORE. As a preliminary, we first recall the well-known concentration results for random matrices, which will be used in the proofs of the error bounds.    

Notation: Let $A,B\in \R^{d\times d}$ be two positive semi-definite (PSD) matrices. We use $A\succeq B\succeq 0$ to denote that $x^\top (A - B)x \geq 0$ for any $x$. As a result, we have $\mathrm{Tr}(A - B)\;\ge\;0$ and $\mathrm{Tr}(A)\;\ge\;\mathrm{Tr}(B)$. 
 
 \begin{theorem}[Two-sided bound on sub-gaussian matrices]\label{thm:bd_subguassian_svd}
 \cite[Theorem 4.6.1]{vershynin2018high} Let $A$ be an $m\times n$ matrix whose rows $A_i$ are independent, mean zero, sub-gaussian isotropic random vectors in $\R^n$. Then for any $t\geq 0$ we have 
 \begin{equation}\label{eq:bound_eigen}
 1-CK^2(\sqrt{\frac{n}{m}}+ \frac {t} {\sqrt{m}} ) \leq s_n(A) \leq s_1(A)\leq   1+ CK^2(\sqrt{\frac{n}{m}}+ \frac {t} {\sqrt{m}} )	
 \end{equation}
with probability at least $1-2e^{t^2/2}$. Here 
$CK^2= \sup_{x: \|x\|=1} \|\langle A_i,x\rangle^2 - 1\|_{\psi_1}$. 
 \end{theorem}

Recall the following Orlicz $\psi_2$-norm for sub-Gaussian distributions and the $\psi_1$-norm for sub-exponential distributions (see, e.g., \cite[Definitions 2.5.6 and 2.7.5]{vershynin2018high}). 
\begin{definition}[ Orlicz $\psi_2$ and $\psi_1$ Norm]
For a real-valued Sub-Gaussian random variable $X$,
\[
  \|X\|_{\psi_2}  : =   \inf\Bigl\{t>0:\;\E\exp\bigl(X^2/t^2\bigr)\le2\Bigr\}.
\]
For a real-valued Sub-Gaussian random vector $X\in\R^n$,
\[
  \|X\|_{\psi_2}  : =\sup_{x\in\R^n: \|x\|=1} \|\langle X,x\rangle\|_{\psi_2} . 
\]
For a real-valued Sub-exponential random variable $Z$,
\[\|Z\|_{\psi_1}  : =   \inf\Bigl\{t>0:\;\E\exp\bigl(|Z|/t\bigr)\le2\Bigr\}. \]
\end{definition}

In particular, when $X\sim \mathcal{N}(0,\sigma_\varepsilon^2 I_n)$, direct computation gives that $Z= X^2-1$ is sub-exponential and 
$$ \|X\|_{\psi_2}\leq \sqrt{8/3}\sigma, \quad \|Z\|_{\psi_1}\leq \frac{2}{e}\sigma_\varepsilon^2.$$
Thus, when the rows $A_i$ are standard Gaussian, we have that \eqref{eq:bound_eigen} holds with $CK^2\leq \frac{2} e< 1$. Thus, Theorem \ref{thm:bd_subguassian_svd} leads to the following bounds on the singular values of matrices with Gaussian rows.  

\begin{corollary}[Two-sided bound on Gaussian matrices]
\label{cor:bd_guassian_svd}
Let $A$ be an $m\times n$ matrix whose rows $A_i$ are independent random vectors with distribution $\mathcal{N}(0,I_n)$.  Then,  for any $t\geq 0$ we have 
 \begin{equation}\label{eq:bound_eigen2}
 1-(\sqrt{\frac{n}{m}}+ \frac {t} {\sqrt{m}} ) \leq s_n(A) \leq s_1(A)\leq   1+ (\sqrt{\frac{n}{m}}+ \frac {t} {\sqrt{m}} )	
 \end{equation}
with probability at least $1-2e^{t^2/2}$. 	
\end{corollary}

\begin{lemma}
Let $X= (x_1,\ldots,x_n)^\top \in\R^{n\times d}$ have i.i.d.\ rows $x_i^\top \sim\mathcal{N}(0,\Sigma_x)$, where the smallest eigenvalue of the covariance $\Sigma_x$ is $\lambda_{\min}(\Sigma_x)>0$. Let $U\in\R^{d\times r_w}$ be column orthonormal with $r_w<n$. Define
\[
  S_s \;=\; \frac1n\,(XU)^\top XU \in\R^{r_w\times r_w}.
\]
Then, for any $0\leq t<\sqrt{n}-\sqrt{r_w}$, with probability at least $1-2\exp(-t^2/2)$ one has
\begin{equation}\label{eq:gram_conc}
  S_s \succeq \lambda_{\min}(\Sigma_x)\;\bigl(1 - \sqrt{r_w/n} - t/\sqrt n\bigr)^2\,I_{r_w}.
\end{equation}
In particular, for any $0\leq  t<1-\sqrt{r_w/n}$, with probability at least $1-2\exp(-n t^2/2)$,
\[
  \lambda_{\min}(\Sigma_x)\;\bigl(1 - \sqrt{r_w/n} - t \bigr)^2
\le \lambda_{\min}(S_s)  \leq
 \lambda_{\max}(\Sigma_x)\;\bigl(1 + \sqrt{r_w/n} + t \bigr)^2
 \]
\end{lemma}

\begin{proof}
Write $X_s = XU\in\R^{n\times r_w}$ and let
\[
  W = X_s\,(\Sigma_x^*)^{-1/2},
  \qquad \Sigma_x^* = U^\top\Sigma_x\,U\in\R^{r_w\times r_w}.
\]
Then the rows of $W$ are i.i.d.\ $\mathcal{N}(0,I_{r_w})$, and
\[
  S_s = \tfrac1n\,X_s^\top X_s
      = (\Sigma_x^*)^{1/2}\,\bigl(\tfrac1nW^\top W\bigr)\,(\Sigma_x^*)^{1/2}.
\]
A standard concentration result for the smallest eigenvalue of Gaussian matrices in Corollary \ref{cor:bd_guassian_svd} states that for any $t\ge0$,
 with probability at least $1-2e^{-t^2/2}$,
\[
  (1-\sqrt{r_w/n}-t/\sqrt n)^2 I_{r_w} \preceq  \tfrac1nW^\top W \preceq    (1+\sqrt{r_w/n}+t/\sqrt n)^2 I_{r_w}. 
\]
Therefore, 
\begin{align*}
 \lambda_{\min}(\Sigma_x^*)\,(1-\sqrt{r_w/n}-t/\sqrt n)^2\,I_{r_w} \preceq & S_s
  = (\Sigma_x^*)^{1/2}\,\tfrac1nW^\top W\,(\Sigma_x^*)^{1/2}  \\
  \preceq & \lambda_{\max}(\Sigma_x^*)\,(1+\sqrt{r_w/n}+t/\sqrt n)^2\,I_{r_w}.
\end{align*}
Since $\lambda_{\min}(\Sigma_x) \le \lambda_{\min}(\Sigma_x^*)\le \lambda_{\max}(\Sigma_x) $, this proves \eqref{eq:gram_conc}.  Replacing $t$ by $\sqrt n\,t$ yields the second statement.
\end{proof}

\begin{proof}[Proof of Proposition \ref{prop:bd_ore_condX}]
By writing $w=U v$ with $ v\in\R^{r_w\times 1}$, the observation model becomes
\[
  Y = XU v + \varepsilon,
  \quad  v\sim\mathcal{N}(0,\Lambda). 
\]
Standard Gaussian linear regression yields the posterior of $v$ is 
\[
   v\mid Y \sim \mathcal{N}(\widehat v,\,\Sigma_v^{\mathrm{post} }),
  \quad
  \Sigma^{\mathrm{post} }_v = \big( \Lambda^{-1} +\tfrac1{\sigma_\varepsilon^2}(XU)^\top XU\big)^{-1}
\]
with posterior mean $\widehat v=(\Lambda^{-1}+\tfrac1{\sigma_\varepsilon^2}(XU)^\top XU)^{-1}(\tfrac1{\sigma_\varepsilon^2}(XU)^\top Y)$.  Hence, the Bayes risk of the ORE $  \widehat{w}^{\mathrm{ORE}}=U\widehat v$ is 
\[
  \mathbb{E}\bigl\|\widehat{w}^{\mathrm{ORE}}-w\bigr\|^2
  = \mathbb{E}\bigl\|\widehat v- v\bigr\|^2
  = \mathbb{E}\bigl[\mathrm{Tr}(\Sigma_v^{\mathrm{post} })\bigr].
\]

By assumption, $\Lambda^{-1} + \tfrac1{\sigma_\varepsilon^2}(XU)^\top XU
    \succeq \tfrac{n\,b 
    }{\sigma_\varepsilon^2}\,I_{r_w}$.
Consequently, 
\[
  \Sigma_v^{\mathrm{post} }
  = \bigl(\Lambda^{-1}+\tfrac1{\sigma_\varepsilon^2}(XU)^\top XU\bigr)^{-1}
  \preceq \tfrac{\sigma_\varepsilon^2}{n\,b
  }\,I_{r_w},\quad  \Rightarrow \mathrm{Tr}(\Sigma_v^{\mathrm{post} })\leq \frac{\sigma_\varepsilon^2\,r_w}{n\,b 
  }. 
\]
Combining the above two estimates, we complete the proof. 
\end{proof}

\begin{proof}[Proof of Proposition \ref{prop:2sideBd_ore}]
By Proposition \ref{prop:bd_ore_condX}, the conditional Bayes-risk given $X$ is bounded by 
$$
\E_{w,\varepsilon}\bigl\|\widehat{w}^{\mathrm{ORE}}-w\bigr\|^2
\;=\;
\mathrm{Tr}\!\bigl(\Sigma_v^{\mathrm{post} }(X)\bigr),
$$
where 
$$
\Sigma_v^{\mathrm{post} }(X)
=\Bigl(\Lambda^{-1}+\tfrac1{\sigma_\varepsilon^2}n S_s \Bigr)^{-1}, \quad 
 S_s \;=\;\frac1n\,(XU)^\top (XU).
$$

\noindent \textbf{Part I: Upper bound.}
Corollary \ref{cor:bd_guassian_svd} implies that for any $t\ge 0$, the minimal eigenvalue of the empirical Gram matrix $S_s$ satisfies  
$$
\mathbb{P}\Bigl(
\lambda_{\min}(S_s)\ge\lambda_{\min}(\Sigma_x)\,b_t
\Bigr)
\;\ge\;
1-2\exp\!\bigl(-nt^2/2\bigr).
$$
Denote $\mathcal A=\{\omega: \lambda_{\min}(S_s(\omega))\ge\lambda_{\min}(\Sigma_x)\,b_t\} $. On $\mathcal A$ we have
$$
\Lambda^{-1} + \tfrac1{\sigma_\varepsilon^2}\,(XU)^\top(XU)
\;\succeq\;
\left( \lambda_1^{-1}+ \tfrac{n}{\sigma_\varepsilon^2}\,\lambda_{\min}(\Sigma_x)\,b_t\right)\,I_{r_w}. 
$$
Thus, $
\Sigma_v^{\mathrm{post} }(X)\;\preceq\;\frac{\sigma_\varepsilon^2}{n\,\lambda_{\min}(\Sigma_x)\,b_t 
}\;I_{r_w},
$ and 
$$
\mathrm{Tr}\!\bigl(\Sigma_v^{\mathrm{post} }(X)\bigr)
\;\le\;
\frac{\sigma_\varepsilon^2\,r_w}{n\,\lambda_{\min}(\Sigma_x)\,b_t 
}.
$$

On $\mathcal A^c$, we have the rough bound by noting that  
$
\Sigma_v^{\mathrm{post} }(X)\;\preceq\; \Lambda, 
$
hence
$$
\mathrm{Tr} \bigl(\Sigma_v^{\mathrm{post} }(X)\bigr) \le \mathrm{Tr}(\Lambda) 
\le  r_w\,\lambda_{\max}(\Lambda).
$$

Putting them all together, in particular,  $\mathbb{P}(\mathcal A^c)\le2e^{-nt^2/2}$, we have 
\begin{align*}
\E_X\Bigl[\mathrm{Tr}(\Sigma_v^{\mathrm{post} }(X))\Bigr]
=& \E\bigl[\mathrm{Tr}(\Sigma_v^{\mathrm{post} }(X))\,1_{\mathcal A}\bigr]
+\E\bigl[\mathrm{Tr}(\Sigma_v^{\mathrm{post} })\,1_{\mathcal A^c}\bigr]	\\
\le &  \frac{\sigma_\varepsilon^2\,r_w}{n\,\lambda_{\min}(\Sigma_x)\,b_t}
\;+\;
r_w\,\lambda_{\max}(\Lambda)\;\cdot2\,e^{-nt^2/2}.
\end{align*}

\noindent \textbf{Part II: lower bound.} 
 To get the lower bound of the Bayes risk, we use the upper tail in the bounds on sub-Gaussian Gram matrices. 
 
First, we find a lower bound for the trace of the posterior covariance conditional on $X$.  
Recall that with probability at least $1-2e^{-nt^2/2}$ one has
$$
S_s \;=\;\frac1n\,(XU)^\top(XU) \;\preceq\;\lambda_{\max}(\Sigma_x)\,a_t\,I_{r_w}, \quad a_t:=\bigl(1+\sqrt{r_w/n}+t\bigr)^2
$$
Denote this event $\mathcal B$. On $\mathcal B$:
$$
\Lambda^{-1}+\frac{1}{\sigma_\varepsilon^2}(XU)^\top(XU)
\;\preceq\;
\Lambda^{-1} \;+\;\frac{n\,\lambda_{\max}(\Sigma_x)a_t}{\sigma_\varepsilon^2}\;I_{r_w}.
$$
Inverting it and using that if $A\preceq B$ then $A^{-1}\succeq B^{-1}$, we have
$$
\Sigma_v^{\mathrm{post} }(X)
=\Bigl(\Lambda^{-1}+\tfrac1{\sigma_\varepsilon^2}(XU)^\top(XU)\Bigr)^{-1}
\;\succeq\;
\Bigl(\Lambda^{-1} + c\,I\Bigr)^{-1},\quad c \;=\;\frac{n\,\lambda_{\max}(\Sigma_x)a_t}{\sigma_\varepsilon^2}.
$$
Hence
$$
\mathrm{Tr}\!\bigl(\Sigma_v^{\mathrm{post} }(X)\bigr)
\;\ge\;
\sum_{i=1}^{r_w}
\frac{1}{\lambda_i^{-1}+c}
\;\ge\;
r_w\;\frac{1}{\lambda_{\min}(\Lambda)^{-1}+c}
\;=\;
\frac{r_w}{\lambda_{\min}(\Lambda)^{-1} + \tfrac{n\,\lambda_{\max}(\Sigma_x)a_t}{\sigma_\varepsilon^2}}.
$$
Rewriting,
$$
\mathrm{Tr}\!\bigl(\Sigma_v^{\mathrm{post} }(X)\bigr)
\;\ge\;
\frac{r_w\,\sigma_\varepsilon^2\,}
     {\sigma_\varepsilon^2  \lambda_{\min}(\Lambda)^{-1} \;+\;n\,\lambda_{\max}(\Sigma_x)\,a_t}.
$$

Second, since $\mathrm{Tr}(\Sigma_v^{\mathrm{post} })\ge0$, taking unconditional expectation, 
 we obtain 
\begin{align*}
\E_X\bigl[\mathrm{Tr}(\Sigma_v^{\mathrm{post} }(X))\bigr]
\ge &
\E\bigl[\mathrm{Tr}(\Sigma_v^{\mathrm{post} })\,1_{\mathcal B}\bigr]
\;\ge\;
\mathbb{P}(\mathcal B)\;\times\;
\frac{r_w\,\sigma_\varepsilon^2\,}
     {\sigma_\varepsilon^2 \lambda_{\min}(\Lambda)^{-1} + n\,\lambda_{\max}\,a_t} \\
   \ge & 
\Bigl(1-2e^{-nt^2/2}\Bigr)\,
\frac{r_w\,\sigma_\varepsilon^2 }
     {\sigma_\varepsilon^2 \lambda_{\min}(\Lambda)^{-1} + n\,\lambda_{\max}(\Sigma_x)\,a_t}\,, 
\end{align*}
where the last inequality follows from $\mathbb{P}(\mathcal B)\ge1-2e^{-nt^2/2}$. 
\end{proof}
	
Lastly, the proof for the covariance of the ORE in Lemma \ref{lemma:cov_ORE} follows from the Law of Total Covariance. 

\begin{proof}[Proof of Lemma \ref{lemma:cov_ORE}]
Since $\widehat{w}^{\mathrm{ORE}}= \E[w \mid X, Y]$, the law of total expectation implies that $\E[\widehat{w}^{\mathrm{ORE}}]= w_0$. 
Applying the Law of Total Covariance to the random vector $w\sim \mathcal{N}(w_0,\Sigma_w)$ and the conditioning variables $(X,Y)$:
\begin{equation}\label{eq:law_total_cov} 
\text{Cov}(w) = \mathbb{E}_{X,Y}[\text{Cov}(w|X,Y)] + \text{Cov}_{X,Y}(\mathbb{E}[w|X,Y]).  
\end{equation}
Note that $\text{Cov}(w)$ is  the prior covariance $\Sigma_w = U\Lambda U^\top$ and $\text{Cov}_{X,Y}(\mathbb{E}[w|X,Y])$ is the covariance of the ORE $\widehat{w}^{\mathrm{ORE}}$ since $\widehat{w}^{\mathrm{ORE}}= \mathbb{E}[w|X,Y]$. Also, note that $\text{Cov}(w|X,Y)$ is the posterior covariance of $w$ given data $(X,Y)$. Writing $w-w_0 = Uv$, we have 
$$
\text{Cov}(w|X,Y) = \text{Cov}(Uv|X,Y) = U \text{Cov}(v|X,Y) U^\top = U \Sigma_v^{\mathrm{post} } U^\top, 
$$ 
where $\Sigma_v^{\mathrm{post} } = \big( \Lambda^{-1} +\tfrac1{\sigma_\varepsilon^2}(XU)^\top XU\big)^{-1} = \sigma^2_\varepsilon \bigl( \sigma^2_\varepsilon \Lambda^{-1} + U^\top X^\top X U \bigr)^{-1}$.

Substituting these into the \eqref{eq:law_total_cov},
$$ \Sigma_w = \mathbb{E}_{X,Y}[U \Sigma_v^{\mathrm{post} } U^\top] + \text{Cov}(\widehat{w}^{\mathrm{ORE}}).
$$
Since $U$ is a constant matrix derived from the prior covariance $\Sigma_w$ and sine $\Sigma_v^{\mathrm{post} }$ does not depend on $Y$, the expectation $\mathbb{E}_{X,Y}[U \Sigma_v^{\mathrm{post} } U^\top]$ simplifies to $U\mathbb{E}_{X}[\Sigma_v^{\mathrm{post} }]U^\top$. Then, the equation for the covariance $\text{Cov}(\widehat{w}^{\mathrm{ORE}})$ follows. 
\end{proof}

\subsection{Additional numerical results}\label{sec:additional_num}

In all experiments, the optimization is performed with the Adam optimizer. We tune the hyperparameters, including the learning rates, decay rates, and regularization parameters, to minimize the validation error (in a relative $L^2$ norm) on a validation dataset with 50 contexts. Experiments are conducted on a single NVIDIA GeForce RTX 3090 GPU with 24 GB memory.

\textbf{Details of experiments in Section 3. } To test the convergence of prior estimates with different training and test task sizes, we have considered $r_w=2$, $\sigma_w=0.01$, $L=4$ layers, and $d_k=10$. As shown in Table \ref{table:3.1}, we have tuned the architecture size to make sure that the transformer is sufficiently expressive to capture the nonlinear mapping from data to $w_\theta$.

\begin{table}[H]
\centering
\caption{Architecture tuning for Section 3}
{\footnotesize\begin{tabular}{|l |c|c|c|c|c|}
\hline
Setting & Para & Train Loss & Ker Err (Train) & Valid Loss & Ker Err (Valid)\\
\hline
\multicolumn{5}{c}{Varying training task sizes and architecture}\\
\hline
ntrain=2500,layer=2,dk=10&12341&0.008548&0.002432&0.327875&0.335600\\
ntrain=2500,layer=4,dk=10&14581&0.008324&0.002960&0.103121&0.103876\\
ntrain=5000,layer=2,dk=10&12341&0.038104&0.036146&0.248010&0.251928\\
ntrain=5000,layer=4,dk=10&14581&0.009714&0.005369&0.019604&0.017697\\
ntrain=10000,layer=2,dk=10&12341&0.081869&0.081732&0.220550&0.226744\\
ntrain=10000,layer=4,dk=10&14581&0.010169&0.005612&0.019404&0.012264\\
ntrain=20000,layer=2,dk=10&12341&0.105982&0.106545&0.118441&0.120079\\
{ntrain=20000,layer=4,dk=10}&14581&0.01084&0.006837&0.010435&0.006819\\
\hline
\multicolumn{5}{c}{Varying training task sizes and architecture}\\
\hline
ntrain=20000,layer=2,dk=10&12341&0.105982&0.106545&0.118441&0.120079\\
ntrain=20000,layer=4,dk=10&14581&0.01084&0.006837&0.010435&0.006819\\
ntrain=20000,layer=6,dk=20&22941&0.007761&0.004146&0.007444&0.003971\\
ntrain=20000,layer=6,dk=40&25181&0.006622&0.004297&0.006191&0.004041\\
ntrain=20000,layer=8,dk=20&27221&0.007751&0.004218&0.007906&0.004721\\
ntrain=20000,layer=8,dk=40&43541&0.006509&0.004235&0.006349&0.004175\\
ntrain=20000,layer=10,dk=20&31501&0.007398&0.004028&0.007225&0.003975\\
ntrain=20000,layer=10,dk=40&51901&0.006675&0.004147&0.006578&0.004015\\
\hline
\end{tabular}}\label{table:3.1}
\end{table}


\textbf{Details of experiments in Section 4.1. }To test the impact from task dimension, we have considered $n_s=20000$, $\sigma_w=0.01$, $L=10$ layers, and $d_k=40$. As shown in Table \ref{table:4.1}, generally a deeper layer and larger $L$ becomes necessary when the task dimension increases.

\begin{table}[H]
\centering
\caption{Test size nTest= 50. Change of intrinsic dimension}
{\footnotesize\begin{tabular}{|l |c|c|c|c|c|}
\hline
Setting & Para & Train Loss & Ker Err (Train) & Valid Loss & Ker Err (Valid)\\
\hline
\multicolumn{5}{c}{$r_w=2$}\\
\hline
ntrain=20000,layer=4,dk=10&14581&0.01084&0.006837&0.010435&0.006819\\
ntrain=20000,layer=6,dk=20&22941&0.007761&0.004146&0.007444&0.003971\\
ntrain=20000,layer=6,dk=40&25181&0.006622&0.004297&0.006191&0.004041\\
ntrain=20000,layer=8,dk=20&27221&0.007751&0.004218&0.007906&0.004721\\
ntrain=20000,layer=8,dk=40&43541&0.006509&0.004235&0.006349&0.004175\\
ntrain=20000,layer=10,dk=20&31501&0.007398&0.004028&0.007225&0.003975\\
ntrain=20000,layer=10,dk=40&51901&0.006675&0.004147&0.006578&0.004015\\
\hline
\multicolumn{5}{c}{$r_w=5$}\\
\hline
ntrain=20000,layer=6,dk=20&22941&0.003881&0.002730&0.003826&0.002852\\
ntrain=20000,layer=6,dk=40&25181&0.002797&0.002159&0.002893&0.002400\\
ntrain=20000,layer=8,dk=20&27221&0.004037&0.002833&0.003837&0.002740\\
ntrain=20000,layer=8,dk=40&43541&0.002946&0.002163&0.002974&0.002197\\
ntrain=20000,layer=10,dk=20&31501&0.003862&0.002667&0.003766&0.002606\\
ntrain=20000,layer=10,dk=40&51901&0.003252&0.002215&0.003219&0.002170\\
\hline
\multicolumn{5}{c}{$r_w=10$}\\
\hline
ntrain=20000,layer=6,dk=20&22941&0.003428&0.003175&0.004089&0.004123\\
ntrain=20000,layer=6,dk=40&25181&0.002221&0.001958&0.002226&0.001926\\
ntrain=20000,layer=8,dk=20&27221&0.003133&0.002887&0.003085&0.002824\\
ntrain=20000,layer=8,dk=40&43541&0.001987&0.001885&0.002126&0.002038\\
ntrain=20000,layer=10,dk=20&31501&0.002700&0.002350&0.002706&0.002432\\
ntrain=20000,layer=10,dk=40&51901&0.002286&0.002204&0.002243&0.002055\\
\hline
\multicolumn{5}{c}{$r_w=20$}\\
\hline
ntrain=20000,layer=6,dk=20&22941&0.017061&0.021798&0.022172&0.028433\\
ntrain=20000,layer=6,dk=40&25181&0.002335&0.002777&0.002463&0.002826\\
ntrain=20000,layer=8,dk=20&27221&0.005263&0.006456&0.006134&0.007667\\
ntrain=20000,layer=8,dk=40&43541&0.001909&0.002360&0.002103&0.002533\\
ntrain=20000,layer=10,dk=20&31501&0.002448&0.002866&0.002716&0.003001\\
ntrain=20000,layer=10,dk=40&51901&0.001673&0.002189&0.001975&0.002431\\
\hline
\end{tabular}}\label{table:4.1}
\end{table}

\textbf{Details of experiments in Section 4.2. } To test the impact from the noise level, we have considered $n_s=20000$, $r_w=2$, $L=10$ layers, and $d_k=40$. For the case of varying the condition number $\kappa(\Sigma_x)$, we fix $n_s=20000$, $r_w=2$, $L=4$ layers, and $d_k=20$. Detailed results are provided in Table \ref{table:4.2}.

\begin{table}[H]
\centering
\caption{Test size nTest= 50. Change of noise}
{\footnotesize\begin{tabular}{|l |c|c|c|c|c|}
\hline
Setting: ntrain=20000,layer=4 & Para & Train Loss & Ker Err (Train) & Valid Loss & Ker Err (Valid)\\
\hline
\multicolumn{5}{c}{Varying noise level}\\
\hline
{dk=10, $\sigma_w=0.01$}&14581&0.01084&0.006837&0.010435&0.006819\\
{dk=10, $\sigma_w=0.05$}&14581&0.044536&0.022718&0.043934&0.025322\\
{dk=10, $\sigma_w=0.1$}&14581&0.083249&0.040972&0.083119&0.045335\\
{dk=10, $\sigma_w=0.5$}&14581&0.324553&0.164703&0.320756&0.165010\\
\hline
\multicolumn{5}{c}{Varying condition number of $\Sigma_x$}\\
\hline
{dk=20, $\kappa(\Sigma_x)=1$}&18661&0.008824&0.003924&0.008882&0.004701\\
{dk=20, $\kappa(\Sigma_x)=1.25$}&18661&0.009738&0.004097&0.009796&0.004158\\
{dk=20, $\kappa(\Sigma_x)=2.5$}&18661&0.012581&0.006706&0.012307&0.005929\\
{dk=20, $\kappa(\Sigma_x)=5$}&18661&0.017430&0.009227&0.017546&0.009204\\
\hline
\end{tabular}}\label{table:4.2}
\end{table}


\end{document}